\documentclass[10pt,twocolumn,letterpaper]{article}

\usepackage{ijcb}
\usepackage{times}
\usepackage{epsfig}
\usepackage{graphicx}
\usepackage{amsmath}
\usepackage{amssymb}
\usepackage{comment}
\usepackage{gensymb}
\usepackage{booktabs, multicol, multirow}
\usepackage{array}
\usepackage{color}
\usepackage{subfigure}
\usepackage{caption}
\usepackage{cite}
\usepackage{balance}
\usepackage{setspace}

\ijcbfinalcopy 

\ifijcbfinal\fi
\newcommand*\samethanks[1][\value{footnote}]{\footnotemark[#1]}

\begin{document}

\title{PF-cpGAN:  Profile to Frontal Coupled GAN for Face Recognition in the Wild}

\author{ Fariborz Taherkhani \thanks{Authors Contributed Equally}, \ Veeru Talreja \samethanks[1], \ Jeremy Dawson, \ Matthew C. Valenti, and \ Nasser M. Nasrabadi\\
West Virginia University\\
{\tt \footnotesize \{ft0009, vtalreja\}@mix.wvu.edu, jeremy.dawson@mail.wvu.edu, valenti@ieee.org, nasser.nasrabadi@mail.wvu.edu}}

\maketitle
\thispagestyle{empty}
\begin{abstract}
   In recent years, due to the emergence of deep learning, face recognition has achieved exceptional success. However, many of these deep face recognition models perform relatively poorly in handling profile faces compared to frontal faces. The major reason for this poor performance is  that it is inherently difficult to learn large pose invariant deep representations that are useful for profile face recognition. In this paper, we hypothesize that the profile face domain possesses a gradual connection with the frontal face domain in the deep feature space. We look to exploit this connection by projecting the profile faces and frontal faces into a common latent space and perform verification or retrieval in the latent domain. We leverage a coupled generative adversarial network (cpGAN) structure to find the hidden relationship between the profile and frontal images in a latent common embedding subspace. Specifically, the  cpGAN framework consists of two GAN-based sub-networks, one dedicated to the frontal domain and the other dedicated to the profile domain. Each sub-network tends to find a projection that maximizes the pair-wise correlation between two feature domains in a common embedding feature subspace.  The efficacy of our approach compared with the state-of-the-art is demonstrated using the CFP, CMU Multi-PIE,  IJB-A, and IJB-C datasets.  
\end{abstract}


\section{Introduction}

Due to the emergence of deep learning, face recognition has achieved exceptional success in recent years \cite{Cao2018PoseRobustFR}. However, many of these deep face recognition models perform relatively poorly in handling profile faces compared to frontal faces \cite{sengupta_wacv_cfp}. In other words, face recognition in the wild, or unconstrained face recognition, is a challenging problem. Pose, expression, and lighting variations are considered to be major obstacles in attaining high unconstrained face recognition performance.  Some methods \cite{Masi_pose_aware_2016,Cao2018PoseRobustFR} attempt to address pose-variation issue by learning pose-invariant features, while some other methods \cite{Yim2015RotatingYF, Yin2017TowardsLF,Cole2017SynthesizingNF,FNM, Tran_disentangled_cvpr_17} try to normalize images (along with identity-preservation) to a single frontal pose before recognition.  However, there are three major difficulties related to face frontalization or normalization in unconstrained environment: 
\begin{itemize}
    \item Complicated face-variations besides pose: In comparison to a controlled environment, there are more complex face variations, e.g., lighting, head pose, expression, in real-world scenarios. It is difficult to directly
warp the input face to a normalized view \cite{FNM}. 
   \item Unpaired data: Undoubtedly, obtaining a strictly normalized face is expensive and time-consuming, but getting an effective pair of target normalized face (i.e., frontal-facing, neutral expression) and an input face is  difficult due to highly imbalanced datasets\cite{FNM}.  
   \item Presence of artifacts:  Synthesized ‘frontal’ faces contain artifacts caused by occlusions and non-rigid expressions.
\end{itemize} 
 
\begin{figure}[t]
\centering
\includegraphics[width=6cm]{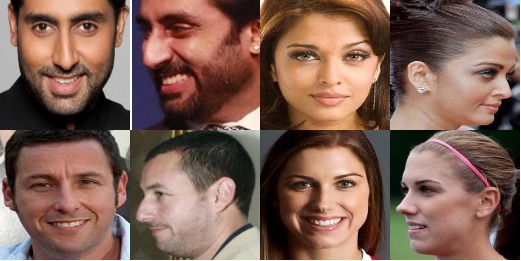}
\caption{Frontal and Profile Images from Celebrity in Frontal Profile (CFP) Dataset.}\label{fig:profile_frontal_images}

\end{figure}
In this paper, we hypothesize that the profile face domain possesses a gradual connection with the frontal face domain in a latent deep feature subspace. We aim to exploit this connection by projecting the profile faces and frontal faces into a common latent subspace and perform verification or retrieval in this latent domain. We propose an embedding model for profile to frontal face verification based on a deep coupled learning framework which uses a generative adversarial network (GAN) to find the hidden relationship between the profile face features and frontal face features in a latent common embedding subspace. 

Our work is conceptually related to the embedding category of super-resolution \cite{shekhar2017synthesis,zhang2015coupled,jiang2016cdmma,Li_2019_LRFRW} in that our approach also performs verification of profile and frontal face in the latent space but not in the image space. From our experiments, we observe that transforming profile and frontal face features to a latent embedding subspace could yield higher performance than image-level face frontalization, which is susceptible to the negative influence of artifacts as a result of image synthesis. To our best knowledge, this study is the first attempt to perform profile-to-frontal face verification in the latent embedding subspace using generative modeling.

This paper makes the following contributions:

\begin{itemize}
    \item A novel profile to frontal face recognition model using coupled GAN framework with multiple loss functions is developed.
    \item Comprehensive experiments using different datasets and a comparison of the proposed method with the state-of-the-art methods have been performed, indicating the efficacy of the proposed GAN framework.
    \item The proposed framework can potentially be used to improve the performance of traditional face recognition methods by integrating it as a preprocessing procedure for face-frontalization schema. 
\end{itemize}


\section{Related Work}
 \textbf{Face recognition using Deep Learning}: Before the advent of deep learning, traditional methods for face recognition (FR) used one or more layer representations, such as the histogram of the feature codes, filtering responses, or distribution of the dictionary atoms \cite{Wang2018DeepFR}.  FR research was concentrated more toward separately improving preprocessing, local descriptors, and feature transformation; however, overall improvement in FR accuracy was very slow. This all changed with the advent of deep learning, and now deep learning is the prominent technique used for FR. 
 
 Recently various deep learning models such as \cite{taherkhani2018deep, Dabouei_2020_WACV} are used as baseline model for FR. Simultaneously, various loss functions have been explored and used in FR. These loss functions can be categorized as the Euclidean-distance-based
loss, angular/cosine-margin-based loss, and softmax loss and its variations. The contrastive loss and the triplet loss are the commonly used Euclidean-distance-based loss functions \cite{Sun_deep_learning_face_representation_nips,schroff2015facenet,taherkhani2019weakly,taherkhani2019matrix}. For avoiding misclassification of difficult samples \cite{talreja2017multibiometric,talreja2018biometrics}, the learned face features need to be well separated. Angular/cosine-margin based loss \cite{Deng_2019_CVPR,Liu2017SphereFaceDH,boumi2020quantifying} are commonly used to make the learned features more separable with a larger angular/cosine distance. Finally, in the category of softmax loss and its variants for  FR \cite{Hasnat2017vonMM,Wang_NormFace_ICM,Liu2017RethinkingFD}, the softmax loss is modified to improve the FR performance as in \cite{Liu2017RethinkingFD}, where the cosine distance among data features is optimized along with normalization of features and weights.

\textbf{Profile-Frontal Face Recognition:}
Face recognition with pose variation in an unconstrained environment is a very challenging problem. Existing methods focus on the pose variation problem by training separate models for learning pose-invariant features \cite{Masi_pose_aware_2016, Cao2018PoseRobustFR}, elaborate
dense 3D facial landmark detection and warping \cite{taigman_deepface}, and  synthesizing a frontal, neutral expression face from a single image \cite{Tran_disentangled_cvpr_17,Yim2015RotatingYF, Yin2017TowardsLF,Cole2017SynthesizingNF,FNM}. For instance, Cao \etal \cite{Cao2018PoseRobustFR} exploit the inherent mapping between profile and frontal faces, and transform a deep profile face representation to a canonical pose by adaptively adding residuals. FF-GAN \cite{Yin2017TowardsLF} solves the problem of large-pose face frontalization in the wild by incorporating a 3D face model into a GAN. Considering photorealistic and identity preserving frontal view synthesis, a domain adaptation strategy for pose invariant face recognition is discussed in \cite{Zhao_PIM_cvpr}. Tran \etal \cite{Tran_disentangled_cvpr_17} propose a GAN framework to rotate the face and disentangle the identity representation by using the pose code. In \cite{FNM}, a face normalization model (FNM) uses a generative adversarial network (GAN) network with 3 distinct losses for generating canonical-view and expression-free frontal images. 

\section {Generative Adversarial Network}

GAN was first introduced by Goodfellow \etal \cite{NIPS_GAN} and has drawn great attention from the deep learning research community due to its remarkable performance on generative tasks. The GAN framework is based on two competing networks --- a generator network G and a discriminator network D. The generator $G(z;\theta_{g})$ is a differentiable function which maps the noise variable $z$ from training noise distribution $p_{z}(z)$ to a data space with distribution $p_{data}$ using the network parameters $\theta_{g}$. On the other hand, the discriminator $D(.;\theta_{d})$ is also a differentiable function, which discriminates between the real data $y$ and  the generated fake data $G(z)$ using a binary classification model. Specifically, the min-max two-player game between the generator and the discriminator provides a simple and powerful way to estimate target distribution and generate novel image samples \cite{FNM}. The loss function $L(D,G)$ for GAN is given as:
 \vspace{-0.25cm}
\begin{equation}\begin{split}
     L(D,G) & = E_{y\sim P_{data}(y)}[\log D(y)]\\ & + E_{z\sim P_{z}(z)}[\log (1-D(G(z)))]. \end{split}
 \end{equation}
 The objective (two player minimax game) for GAN is as:
 \begin{equation}\begin{split}
     \min_{G}\max_{D} L(D,G) & =\min_{G}\max_{D}[E_{y\sim P_{data}(y)}[\log D(y)]\\ & + E_{z\sim P_{z}(z)}[\log (1-D(G(z)))]].\end{split}\label{eq:2}
 \end{equation}
 
 Another variant of GAN is the Conditional GAN, which was introduced by Mirza and Osindero \cite{mirza_2014_conditional}. In conditional GAN, both the generator and  discriminator are conditioned on an additional variable $x$. This additional variable could be any kind of auxiliary information such as discrete labels \cite{mirza_2014_conditional}, and text \cite{RAYLLS_16}. The loss function for the conditional GAN is given as:
 
 \begin{equation}\begin{split}
     L_{c}(D,G) & = E_{y\sim P_{data}(y)}[\log D(y|x)]\\ & + E_{z\sim P_{z}(z)}[\log (1-D(G(z|x)))].\end{split}\label{eq:3}
 \end{equation} Hereafter, we will denote the objective for the conditional GAN as $F_{cGAN}(D,G,y,x)$, which is given by: \begin{equation}\begin{split}
F_{cGAN}(D,G,y,x) & = \min_{G}\max_{D} [E_{y\sim P_{data}(y)}[\log D(y|x)]\\ & + E_{z\sim P_{z}(z)}[\log (1-D(G(z|x)))]].\end{split}\label{eq:4}
 \end{equation}

\section{Proposed Method}
Here, we describe our method for profile to frontal face recognition. In contrast to the face normalization methods, we do not perform pose normalization (i.e., frontalization) on each profile  image  before matching. Instead, we seek to project the profile and frontal face images to a common latent low-dimensional embedding subspace using generative modeling. Inspired by the success of GANs \cite{NIPS_GAN}, we explore adversarial networks to project profile and frontal images to a common subspace for recognition. 
 
\begin{figure}[t]
\centering
\includegraphics[width=8.1cm]{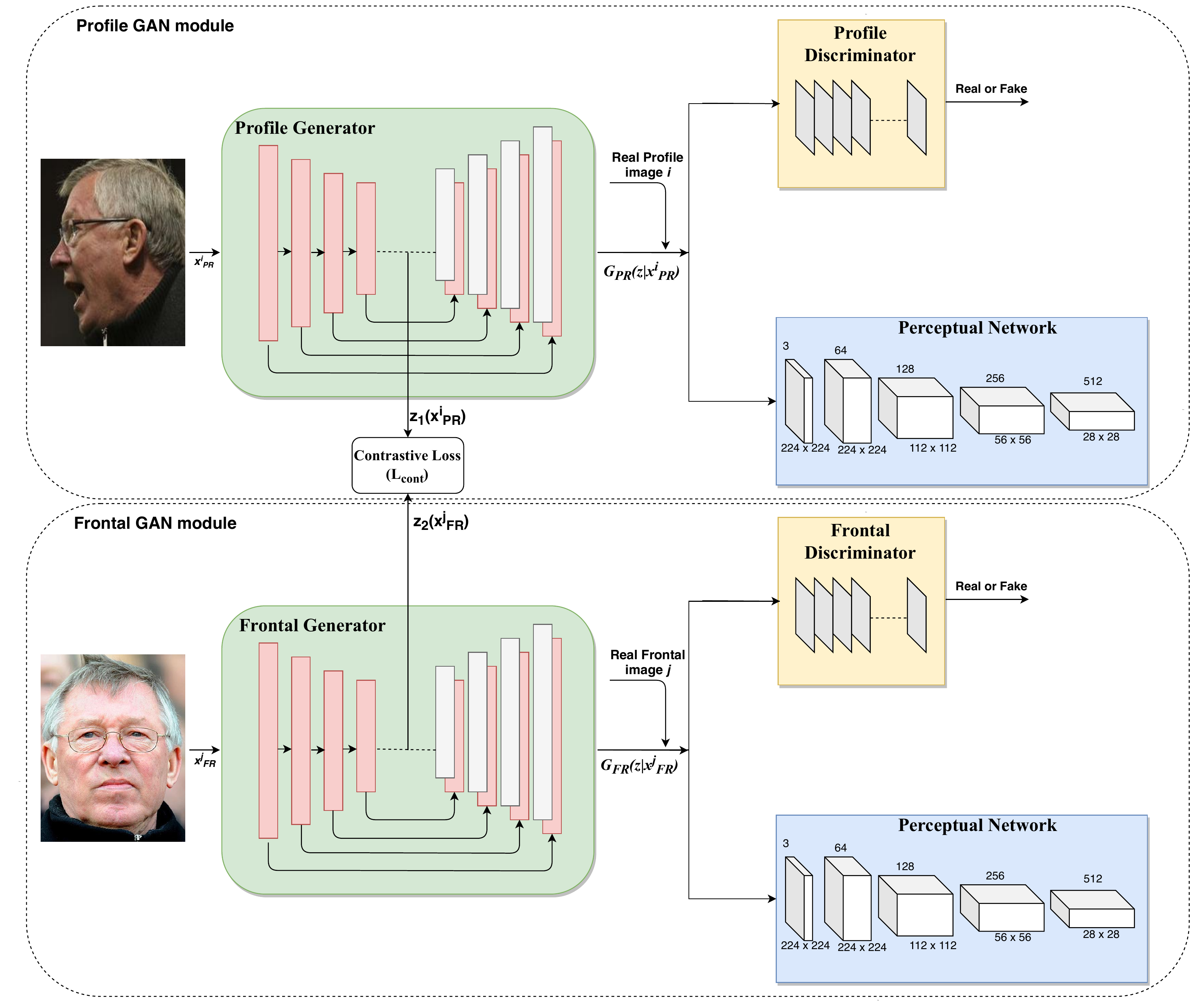}
\caption{Block diagram of PF-cpGAN.}\label{fig:arch}
\vspace{-5.5mm}
\end{figure}

 The framework of proposed profile to frontal coupled generative adversarial  network (PF-cpGAN; shown in Fig. \ref{fig:arch}) consists of two modules, where each module contains a GAN architecture made of a generator and a discriminator. The generators that we have used in both  modules are U-net auto-encoders that are coupled together using a contrastive loss function. In addition to  adversarial, and contrastive loss, we propose to guide the generators using a perceptual loss \cite{Johnson2016PerceptualLF} based on the VGG 16 architecture, as well as an $L_2$ reconstruction error. The perceptual loss helps in sharp and realistic reconstruction of the images. 
 
 \subsection{Profile to Frontal Coupled GAN}
 
 The main objective of PF-cpGAN is the recognition of profile face images with respect to a gallery of frontal face images, which have not been seen during the training. The matching of the profile and the frontal face images is performed in a common embedding subspace. PF-cpGAN consists of two modules: a profile GAN module and a frontal GAN module, both consisting of a GAN (generator + discriminator), and a perceptual network based on VGG-16.
 

 For the generators, we  use a U-Net \cite{ronneberger_2015_unet} auto-encoder architecture. The primary reason for using U-Net
is that the encoder-decoder structure tends to extract global features and generates images by leveraging this overall information, which is very useful for global shape transformation
tasks such as profile to frontal image conversion. Moreover, for many image translation problems, there is a significant amount of low-level information that needs to be shared between the input and output, and it is desirable to pass this information directly across all the layers including the bottleneck. Thus, using skip-connections, as in U-net, provides a means for the encoder-decoder structure to circumvent the bottleneck and pass the information over to other layers. 

For discriminators, we have used patch-based discriminators \cite{Isola_2017_ImagetoImageTW}, which are trained iteratively along with the respective generators. $L_1$ loss performs very well when trying to preserve the low frequency details but fails to preserve the high-frequency details. However, using a patch-based discriminator that penalizes structure at the scale of the patches, ensures the preservation of high-frequency details, which are usually lost when only $L_1$ loss is used. 

The final objective of  PF-cpGAN is to find the  hidden  relationship  between  the  profile face  features and frontal face features in a latent common embedding subspace. To find this common subspace between the two domains, we couple the two generators via a contrastive loss function, $L_{cont}$. 
 
 This loss function  $(L_{cont})$ is a distance-based loss function, which tries to ensure that semantically similar examples (genuine pairs i.e., a profile image of a subject with its corresponding frontal image) are embedded closely in the common embedding subspace, and, simultaneously, semantic dissimilar examples (impostor pairs i.e., a profile image of a subject and a frontal image of a different subject) are pushed away from each other in the common embedding subspace. The contrastive loss function is defined as:

 \begin{equation}
\begin{split}
L_{cont}(z_1&(x^i_{PR}),z_2(x^j_{FR}),Y)= \\ & 
  (1-Y)\frac{1}{2}(D_z)^2 + (Y)\frac{1}{2}(\mbox{max}(0,m-D_z))^2,  
  \end{split}
  \end{equation}where
 $x^i_{PR}$ and $x^j_{FR}$ denote the input profile and frontal face image, respectively. The variable $Y$ is a binary label, which is equal to 0 if $x^i_{PR}$ and $x^j_{FR}$ belong to the same class (i.e., genuine pair), and equal to 1 if $x^i_{PR}$ and $x^j_{FR}$ belong to the different class (i.e., impostor pair). $z_1(.)$ and $z_2(.)$ denote only the encoding functions of the U-Net auto-encoder to transform  $x^i_{PR}$ and $x^j_{FR}$, respectively into a common latent embedding subspace. The value $m$ is the contrastive margin and is used to ``tighten" the constraint. $D_z$ denotes the Euclidean distance between the outputs of the functions $z_1(x^i_{PR})$ and $z_2(x^j_{FR})$.
 
 \begin{equation}
     D_z=\left\lVert z_1(x^i_{PR})-z_2(x^j_{FR})\right\rVert_2.
 \end{equation}
Therefore, if $Y=0$ (i.e., genuine pair), then the contrastive loss function $(L_{cont})$ is given as:
 \begin{equation}
L_{cont}(z_1(x^i_{PR}),z_2(x^j_{FR}),Y)  = \frac{1}{2}\left\lVert z_1(x^i_{PR})-z_2(x^j_{FR})\right\rVert^2_2, 
\end{equation}

and if $Y=1$ (i.e., impostor pair), then contrastive loss function $(L_{cont})$ is :
  \begin{equation}
  \begin{split}
L_{cont}(z_1(x^i_{PR}),&z_2(x^j_{FR}),Y)  = \\ & \frac{1}{2}\mbox{max}\biggl(0,m-\left\lVert z_1(x^i_{PR})-z_2(x^j_{FR})\right\rVert^2_2\biggr).
\end{split}
\end{equation}

 Thus, the total loss  for coupling the profile generator and the frontal generator  is denoted by $L_{cpl}$ and is given as: 



\vspace{-0.25cm}
\begin{equation}
    \begin{split}
        L_{cpl}=\frac{1}{N^2}\sum_{i=1}^{N}\sum_{j=1}^{N}L_{cont}(z_1(x^i_{PR}),z_2(x^j_{FR}),Y), 
    \end{split}\label{eq:6}
\end{equation}
where N is the number of training samples. The contrastive loss in the above equation can also be replaced by some other distance-based metric, such as the Euclidean distance. However, the main aim of using the contrastive loss is to be able to use the class labels implicitly and find the discriminative embedding subspace, which may not be the case with some other metric such as the Euclidean distance. This discriminative embedding subspace would be useful for matching of the profile images with the frontal images.
\subsection{Generative Adversarial Loss}

Let the generators (profile generator and frontal generator) that reconstruct the corresponding profile and frontal image from the input profile  and frontal image, be denoted as $G_{PR}$ and $G_{FR}$, respectively. The patch-based discriminators used for the profile and frontal GANs are denoted as $D_{PR}$ and $D_{FR}$. For the proposed method, we have used the conditional GAN, where  the generator networks $G_{PR}$ and $G_{FR}$  are conditioned on input profile and frontal face images, respectively.  We have used the conditional GAN loss function \cite{mirza_2014_conditional} to train the generators and the corresponding discriminators in order to ensure that the discriminators cannot distinguish the  images reconstructed by the generators from the corresponding ground truth images. Let $L_{PR}$ and $L_{FR}$ denote the conditional GAN loss functions for the profile and the frontal GANs, respectively, where $L_{PR}$ and $L_{FR}$ are given as: 

\begin{equation}
L_{PR}=F_{cGAN}(D_{PR},G_{PR},y^i_{PR},x^i_{PR}),
\end{equation}
\begin{equation}
L_{FR}=F_{cGAN}(D_{FR},G_{FR},y^j_{FR},x^j_{FR}), 
\end{equation}where function $F_{cGAN}$ is the conditional GAN objective defined in (\ref{eq:4}). The term $x^i_{PR}$ denotes the profile image used as a condition for the profile GAN, and $y^i_{PR}$  denotes the real profile image.  Note that the real profile image $y^i_{PR}$ and the network condition given by $x^i_{PR}$ are the same. Similarly, $x^j_{FR}$ denotes the frontal image used as a condition for the frontal GAN and $y^j_{FR}$  denotes the real frontal image. Again, the real frontal image $y^j_{FR}$ and the network condition given by $x^j_{FR}$ are the same. The total loss for the coupled conditional GAN is given by:


\begin{equation}
L_{GAN}=L_{PR}+L_{FR}.
\end{equation}




\subsection{$L_2$ Reconstruction Loss}

We also consider the $L_2$ reconstruction loss for both the profile GAN and frontal GAN. The $L_2$ reconstruction loss measures the reconstruction error in terms of the Euclidean distance between the reconstructed image and the corresponding real image. Let $L_{2_{PR}}$ denote the reconstruction loss for the profile GAN and is defined as: 
\begin{equation}
    L_{2_{PR}}=\left\lVert G_{PR}(z|x^i_{PR})-y^i_{PR}\right\rVert^2_2,
\end{equation}where $y^i_{PR}$ is the ground truth profile image, $G_{PR}(z|x^i_{PR})$ is the output of the profile generator.

Similarly, Let $L_{2_{FR}}$ denote the reconstruction loss for the frontal GAN: 

\begin{equation}
    L_{2_{FR}}=\left\lVert G_{FR}(z|x^j_{FR})-y^j_{FR}\right\rVert^2_2,
\end{equation}where $y^j_{FR}$ is the ground truth frontal image, $G_{FR}(z|x^j_{FR})$ is the output of the frontal generator.

The total $L_2$ reconstruction loss function is given by:
\begin{equation}
    L_{2}=\frac{1}{N^2}\sum_{i=1}^{N}\sum_{j=1}^{N}(L_{2_{PR}}+L_{2_{FR}}).
\end{equation}

\subsection{Perceptual Loss}\label{subsec:percloss}
In addition to the GAN loss and the reconstruction loss which are used to guide the generators, we have also used perceptual loss, which was introduced in \cite{Johnson2016PerceptualLF} for style transfer and super-resolution. The perceptual loss function is used to compare high level differences, like content and style discrepancies, between images. The perceptual loss function involves comparing two images based on high-level representations from a pretrained CNN, such as VGG-16 \cite{simonyan2014very}. The perceptual loss function is a good alternative to solely using  $L_1$ or $L_2$ reconstruction error, as it gives better and sharper high quality reconstruction images \cite{Johnson2016PerceptualLF}.    

In our proposed approach, perceptual loss is added to both the profile and the frontal module using a pre-trained VGG-16 \cite{simonyan2014very} network. We extract the high-level features (ReLU3-3 layer) of VGG-16 for both the real input image and the reconstructed output of the U-Net generator. The $L_1$ distance between these features of real and reconstructed images is used to guide the generators $G_{PR}$ and $G_{FR}$ . The perceptual loss for profile network is defined as:

\begin{equation}
    \begin{split}
        L_{P_{PR}}=&\frac{1}{C_pW_pH_p}\sum_{c=1}^{C_{p}}\sum_{w=1}^{W_{p}}\sum_{h=1}^{H_{p}} \\ & \left\lVert V(G_{PR}(z|x^i_{PR}))^{c,w,h}-V(y^i_{PR})^{c,w,h}\right\rVert,
    \end{split}
\end{equation}
where $V(.)$ denotes a particular layer of the VGG-16, where the layer dimensions are given by $C_p$, $W_p$, and $H_p$.

Likewise the perceptual loss for frontal network is:

\begin{equation}
    \begin{split}
        L_{P_{FR}}=&\frac{1}{C_pW_pH_p}\sum_{c=1}^{C_{p}}\sum_{w=1}^{W_{p}}\sum_{h=1}^{H_{p}} \\ & \left\lVert V(G_{FR}(z|x^j_{FR}))^{c,w,h}-V(y^j_{FR})^{c,w,h}\right\rVert.
    \end{split}
\end{equation}

The total perceptual loss function is given by:
\begin{equation}
    L_{P}=\frac{1}{N^2}\sum_{i=1}^{N}\sum_{j=1}^{N}(L_{P_{PR}}+L_{P_{FR}}).
\end{equation}

\subsection{Overall Objective Function}
The overall objective function for learning the network parameters in the proposed method is given as the sum of all the  loss functions defined above:
\vspace{-0.25cm}
\begin{equation}
    \begin{split}
       L_{tot}=L_{cpl}+ \lambda_1 L_{GAN} + \lambda_2 L_{P}+ \lambda_3 L_2,
    \end{split}\label{eq:20}
\end{equation}
where $L_{cpl}$ is the coupling loss, $L_{GAN}$ is the total generative adversarial loss, $L_{P}$ is the total perceptual loss, and $L_2$ is the total reconstruction error. Variables $\lambda_1, \lambda_2,$ and $\lambda_3$ are the hyper-parameters to weigh the different loss terms. 

\section{Experiments}

 We initially describe our training setup and the datasets that we have used in our experiments. We show the efficiency of our method for the task of frontal to profile face verification by comparing its performance with state-of the-art face verification  methods  across pose-variation. Moreover,  we explore the effect of face yaw in our  algorithm. Finally, we conduct an ablation study to investigate the effect of each term in our total training loss defined in (\ref{eq:20}).

\subsection{Experimental Details}

\textbf{Datasets}: The Celebrities in Frontal-Profile (CFP) dataset \cite{sengupta_wacv_cfp}  is a mixture of constrained (i.e., carefully collected under different pose, illumination and expression conditions) and unconstrained (i.e., collected images from the Internet) settings. CFP includes 500 celebrities, averaging ten frontal and four profile face images per each celebrity. Following the standard 10-fold protocol \cite{sengupta_wacv_cfp}, we divide the dataset into 10 folds, each of which consists of 350 same and 350 different pairs generated from 50 subjects (i.e., 7 same and 7 different  pairs for each subject). 

The CMU Multi-PIE database \cite{CMU-PIE} contains 750,000 images of 337 subjects. Subjects were imaged from  15 viewing angles and 19 illumination conditions while exhibiting a range of facial expressions. It is the largest database for graded evaluation with respect to pose, illumination, and expression variations. For fair comparison, the database setting was made consistent with FNM \cite{FNM}, where  250 subjects from Multi-PIE have been used. The training set and testing split is consistent with FNM.

The IARPA Janus Benchmark A (IJB-A) \cite{IJB-A} is a  challenging dataset collected under complete unconstrained conditions covering full pose variation (yaw angles $-90\degree$ to $+90\degree$). IJB-A contains 500 subjects with 5,712 images and 20,414 frames extracted from videos. Following the standard protocol in \cite{IJB-A}, we evaluate our method on both verification and identification. The IARPA Janus Benchmark B (IJB-B) dataset \cite{IJB-B} builds on the IJB-A  by adding more 1345 subjects making it a total of 1845 subjects, and a total of 21,798 still images and 55,026 frames from 7,011 videos. The IARPA Janus Benchmark C (IJB-C) dataset \cite{IJB-C} builds on IJB-A, and IJB-B datasets and has a total of 31,334 images for a total number of 3,531 subjects. We have also evaluated our method on IJB-A and IJB-C datasets.   

\textbf{Implementation Details}: We have implemented the U-Net with ResNet-18 \cite{He2015_Resnet} encoder pre-trained on ImageNet. We have added an additional fully-connected layer after the average pooling layer for ResNet-18 for our U-Net encoder. The U-Net decoder has the same number of layers as the encoder. The entire framework has been implemented in Pytorch. For convergence,  $\lambda_1$ is set to 1, and $\lambda_2$, and $\lambda_3$ are both set to 0.25. We used a batch size of 128 and an Adam optimizer \cite{Kingma2015AdamAM} with first-order momentum of 0.5, and learning rate of 0.0004. We have used the ReLU activation function for the generator and Leaky ReLU with a slope of 0.3 for the discriminator.  


For training,  genuine and impostor pairs were required. The genuine/impostor pairs are created by frontal and profile images of the same/different subject. During the experiments, we ensure that  the training set are balanced  by using the same number of genuine and impostor pairs. 

\begin{table}[t]
\centering

\caption{Performance comparison on CFP dataset. Mean Accuracy and equal error rate (EER) with standard deviation over 10 folds.}
\scalebox{0.75}{\begin{tabular}{c c c c c}
 \hline
 &\multicolumn{2}{c}{Frontal-Profile} &\multicolumn{2}{c}{Frontal-Frontal}\\ [0.5ex] 
 \hline
 Algorithm&Accuracy&EER&Accuracy&EER  \\ \hline
 HoG+Sub-SML \cite{sengupta_wacv_cfp} &77.31(1.61)&22.20(1.18)&88.34(1.31)&11.45(1.35) \\ \hline
  LBP+Sub-SML \cite{sengupta_wacv_cfp} &70.02(2.14)&29.60(2.11)&83.54(2.40)&16.00(1.74) \\ \hline
   FV+Sub-SML \cite{sengupta_wacv_cfp} &80.63(2.12)&19.28(1.60)&91.30(0.85)&8.85(0.74) \\ \hline
    FV+DML \cite{sengupta_wacv_cfp}  &58.47(3.51)&38.54(1.59)&91.18(1.34)&8.62(1.19) \\ \hline
    Deep Features \cite{deep_features} &84.91(1.82)&14.97(1.98)&96.40(0.69)&3.48(0.67) \\ \hline
    PR-REM \cite{Cao2018PoseRobustFR} &93.25(2.23)&7.92(0.98)&98.10(2.19)&1.10(0.22) \\ \hline
    PF-cpGAN&93.78(2.46)&7.21(0.65)&98.88(1.56)&0.93(0.14) \\ \hline

\end{tabular}}
\label{table:table_cfp}
\end{table}

\subsection{Evaluation on CFP with Frontal-Profile Setting}

We first perform evaluation on the Celebrities in Frontal-Profile (CFP) dataset\cite{sengupta_wacv_cfp}, a challenging dataset created to examine the problem of frontal to profile face verification in the wild. We follow the standard 10-fold protocol\cite{sengupta_wacv_cfp} in our evaluation. The same protocol is applied on both the Frontal-Profile and Frontal-Frontal settings. For fair comparison and as given in \cite{sengupta_wacv_cfp}, we consider different types of feature extraction techniques like HoG\cite{hog}, LBP\cite{lbp}, and Fisher Vector\cite{fisher_vector_dml} along with metric learning techniques like Sub-SML\cite{sub_sml}, and Diagonal metric learning (DML) as reported in \cite{fisher_vector_dml}. We also compare against deep learning techniques, including Deep Features\cite{deep_features}, and PR-REM\cite{Cao2018PoseRobustFR}. The results are summarized in Table \ref{table:table_cfp}. 

We can observe from Table \ref{table:table_cfp} that our proposed framework, PF-cpGAN, gives much better performance than the methods that use standard hand-crafted features of HoG, LBP, or FV, providing  minimum of $13\%$ improvement in accuracy with a $12\%$ decrease in EER for the profile-frontal setting. PF-cpGAN also improves on the performance of the Deep Features by approximately $9\%$  with a $7.5\%$ decrease in EER for the profile-frontal setting. Finally, PF-cpGAN performs on-par with the best deep learning method of PR-REM, and, in-fact, does slightly better than  PR-REM by  $\approx 0.5\%$ improvement in accuracy with a $0.7\%$ decrease in EER for the profile-frontal setting. This performance improvement clearly shows that usage of a GAN framework for projecting the profile and frontal images in the latent embedding subspace and maintaining the sematic similarity in the latent space is better than some other deep learning techniques such as Deep Features or PR-REM.

\begin{table}[t]
\centering

\caption{Performance comparison on IJB-A benchmark. Results reported are the 'average$\pm$standard deviation' over the 10 folds specified in the IJB-A protocol. Symbol '-' indicates that the metric is not available for that protocol.}
\scalebox{0.65}{\begin{tabular}{c c c c c}
 \hline
\multicolumn{1}{c}{\multirow{2}{*}{Method}} &\multicolumn{2}{c}{Verification} &\multicolumn{2}{c}{Identification}\\ [0.5ex] 
 \cline{2-5}
 &GAR@ FAR$=0.01$&GAR@ FAR$=0.001$&@ Rank-1 &@ Rank-5  \\ \hline \hline
 OPENBR \cite{klare2015pushing} & $23.6\pm0.9$&$10.4\pm1.4$&$24.6\pm1.1$&$37.5\pm0.8$\\
 GOTS \cite{klare2015pushing} &$40.6\pm1.4$&$19.8\pm0.8$&$43.3\pm2.1$&$59.5\pm2.0$ \\
 PAM \cite{Masi_pose_aware_2016} &$73.3\pm1.8$&$55.2\pm3.2$&$77.1\pm1.6$&$88.7\pm0.9$ \\
 DCNN \cite{chen2016unconstrained} &$78.7\pm4.3$&-&$85.2\pm1.8$&$93.7\pm1.0$ \\
 DR-GAN \cite{tran2017disentangled} &$77.4\pm2.7$&$53.9\pm4.3$&$85.5\pm1.5$&$94.7\pm1.1$ \\
 FF-GAN \cite{yin2017towards} &$85.2\pm1.0$&$66.3\pm3.3$&$90.2\pm0.6$&$95.4\pm0.5$ \\
 FNM \cite{FNM} &$93.4\pm0.9$&$83.8\pm2.6$&$96.0\pm0.5$&$98.6\pm0.3$ \\
 PR-REM \cite{Cao2018PoseRobustFR} &$94.4\pm0.9$&$86.8\pm1.5$&$94.6\pm1.1$&$96.8\pm1.0$ \\
 PF-cpGAN&$95.8\pm0.82$&$91.2\pm1.3$&$97.6\pm1.0$&$98.8\pm0.4$ \\

\end{tabular}}
\label{table:table_ijb}
\end{table}

\begin{figure*}[h]
\centering     
\subfigure[ IJBA]{\label{fig:a}\includegraphics[scale=0.465]{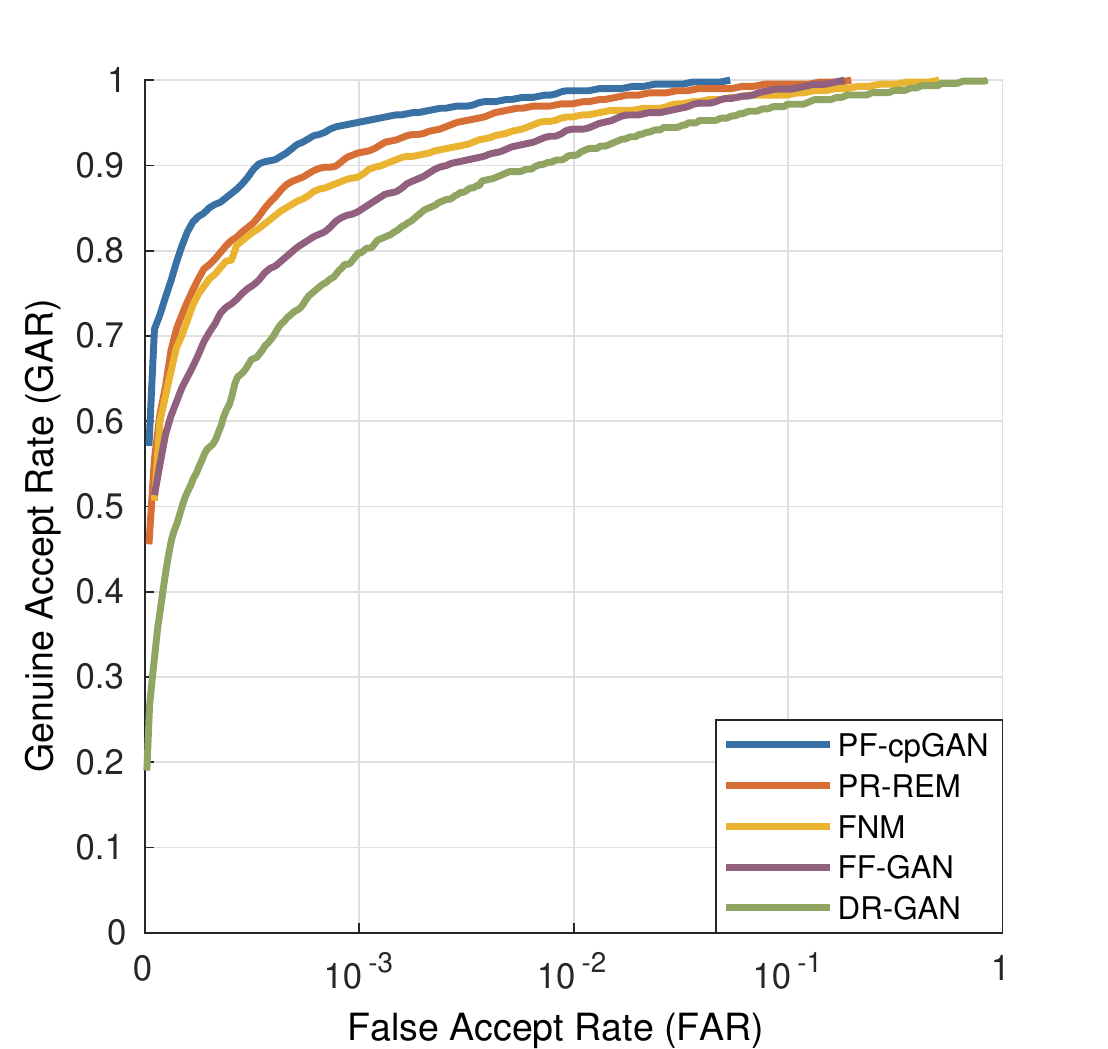}}
\subfigure[CMU Multi-PIE]{\label{fig:b}\includegraphics[scale=0.46]{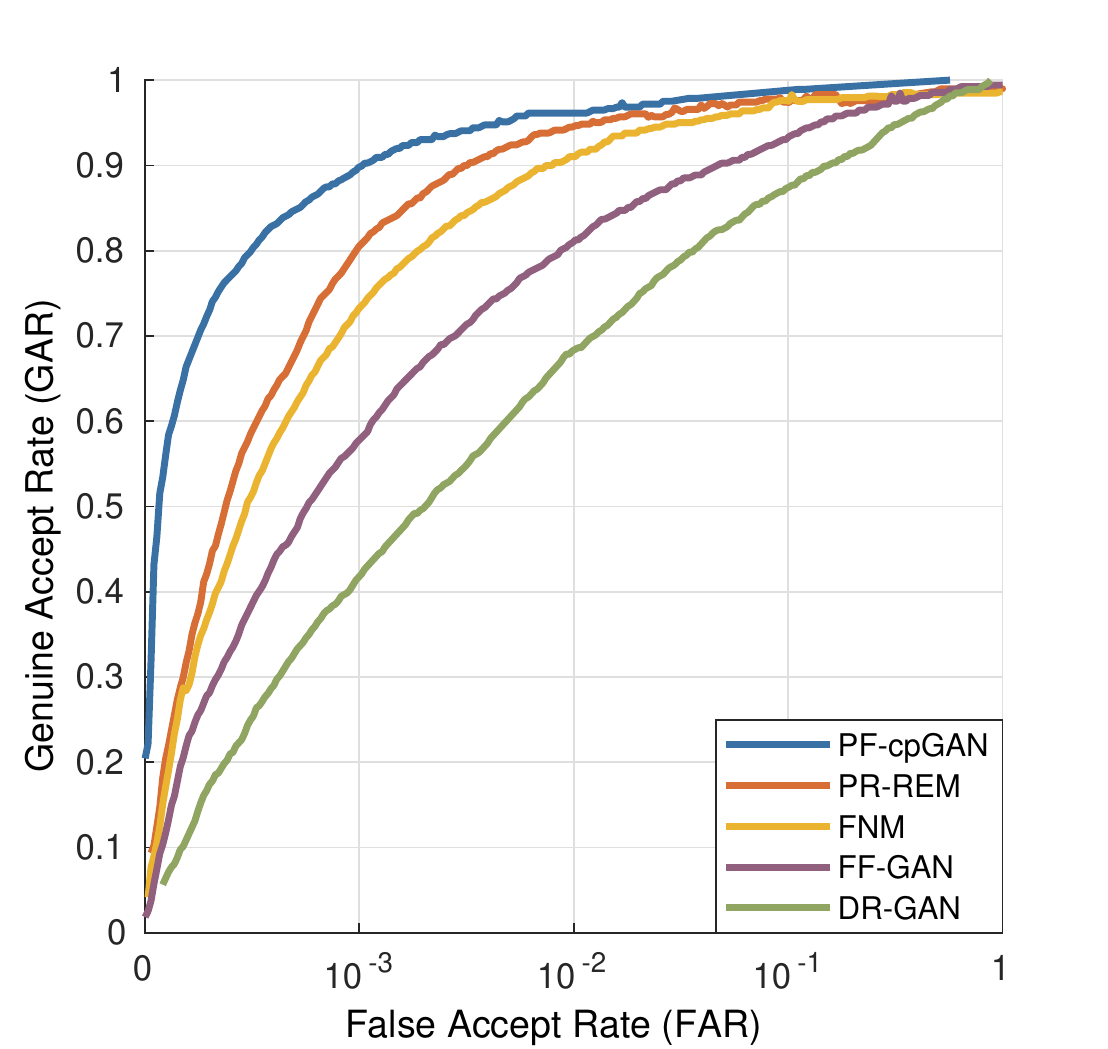}}
\subfigure[Ablation Study]{\label{fig:ablation}\includegraphics[scale=0.46]{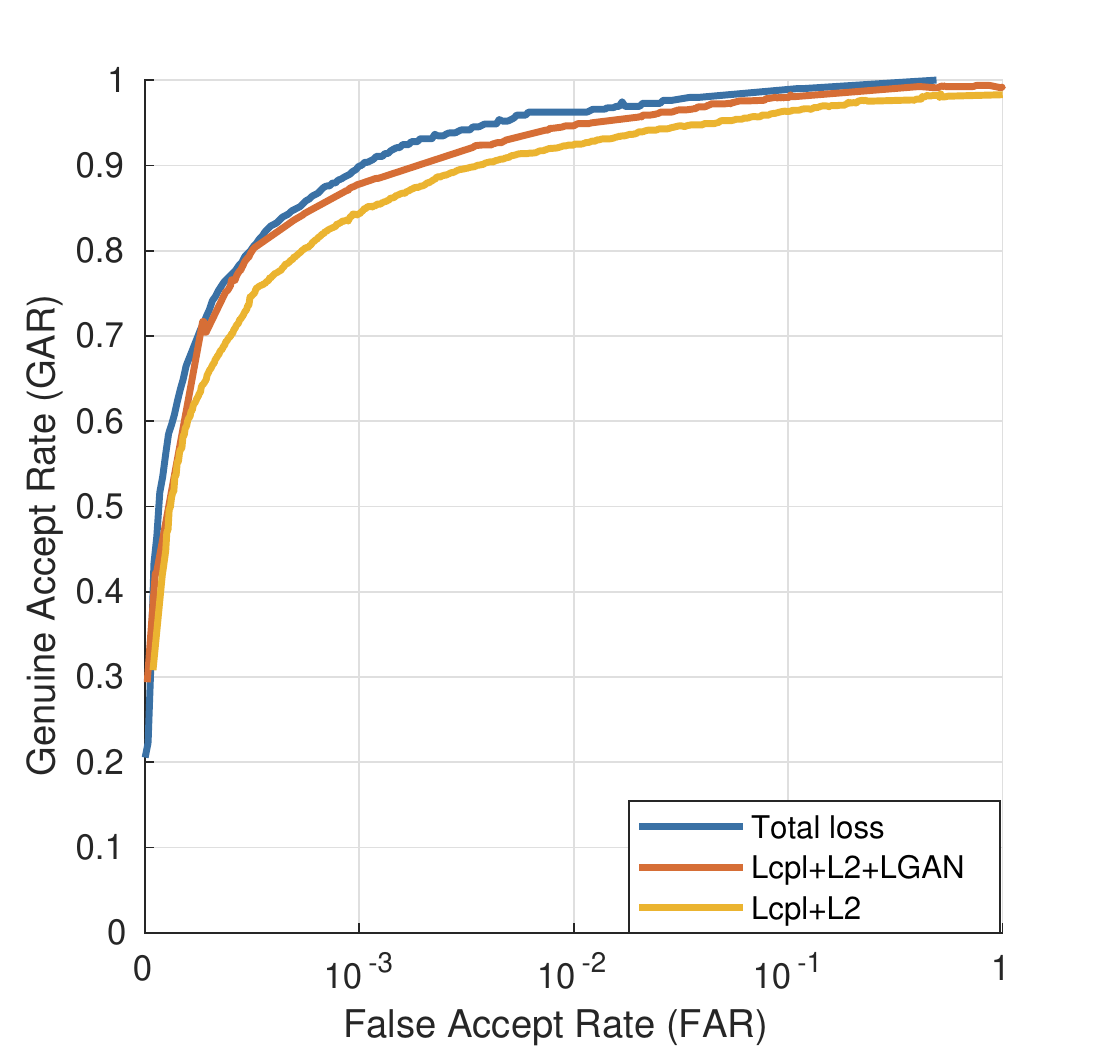}}
\vspace{-0.40cm}
\caption{ROC curve comparison against the baselines for different datasets is shown in (a) and (b). In (c), we show the ROC curves showing the importance of different loss functions for ablation study.}
\label{fig:ROC}
\vspace{-1mm}
\end{figure*}

\subsection{Evaluation on IJB-A and IJB-C}
\begin{figure*}[t]
\centering
\includegraphics[scale=0.16]{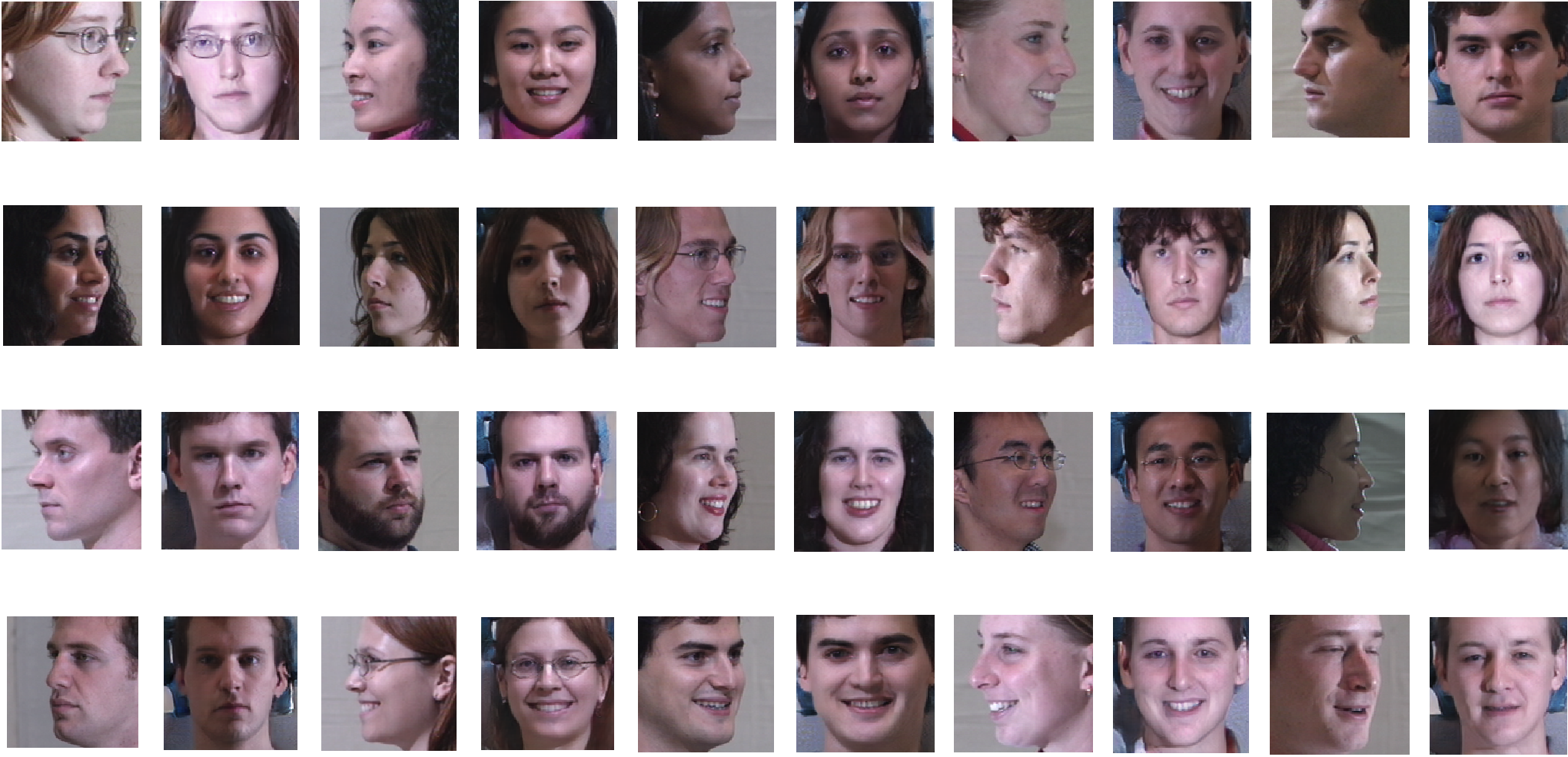}
\caption{Reconstruction of frontal images at the output of the  frontal U-Net generator with profile images as input to the profile U-Net generator. Every odd number column represent the input profile image and every even number column represents the output frontal image. The input images belong to the CMU-MultiPIE dataset. }\label{fig:profile_frontal_images}
\vspace{-1mm}
\end{figure*}
Here, we focus on unconstrained face recognition on IJB-A dataset to quantify the superiority of our PF-cpGAN for profile to frontal face recognition. Some of the baselines for comparison on IJB-A are DR-GAN \cite{Tran_disentangled_cvpr_17}, FNM \cite{FNM}, PR-REM \cite{Cao2018PoseRobustFR},and FF-GAN \cite{Yin2017TowardsLF}. We have also compared them with other methods as listed in \cite{FNM} and shown in Table \ref{table:table_ijb}. As shown in Table \ref{table:table_ijb}, we perform better than the state-of-the-art methods for both verification and identification. Specifically, for verification, we improve the genuine accept rate (GAR) by at least $1.4\%$ compared to other
methods. For instance, at the false accept rate (FAR) of 0.01, the best previously-used method is PR-REM, with an average GAR of $94.4\%$. PF-cpGAN improves upon PR-REM and gives an average GAR of $95.8\%$ at the same FAR.  We also show improvement on  identification. Specifically, the rank-1 recognition rate shows an improvement of around $1.6\%$ in comparison to the best state-of-the-art method, FNM \cite{FNM}.

\begin{table}[t]
\centering

\caption{Performance comparison on IJB-C benchmark. Results reported are the 'average$\pm$standard deviation' over the 10 folds specified in the IJB-C protocol. Symbol '-' indicates that the metric is not available for that protocol.}
\scalebox{0.65}{\begin{tabular}{c c c c c}
 \hline
\multicolumn{1}{c}{\multirow{2}{*}{Method}} &\multicolumn{2}{c}{Verification} &\multicolumn{2}{c}{Identification}\\ [0.5ex] 
 \cline{2-5}
 &GAR@ FAR$=0.01$&GAR@ FAR$=0.001$&@ Rank-1 &@ Rank-5  \\ \hline \hline
 GOTS \cite{IJB-C} &$62.1\pm1.1$&$36.3\pm1.2$&$38.5\pm1.6$&$53.8\pm1.8$ \\
 FaceNet \cite{facenet}
 &$82.3\pm1.18$&$66.3\pm1.3$&$70.4\pm1.2$&$78.8\pm2.3$ \\
 VGG-CNN \cite{Vgg_CNN}
 &$87.2\pm1.09$&$74.3\pm0.9$&$79.6\pm1.04$&$87.8\pm1.3$ \\
 FNM \cite{FNM} &$91.2\pm0.8$&$80.4\pm1.8$&$84.6\pm0.6$&$93.7\pm0.9$ \\
 PR-REM \cite{Cao2018PoseRobustFR} &$92.1\pm0.8$&$83.4\pm1.5$&$83.1\pm0.4$&$92.6\pm1.1$\\
 PF-cpGAN&$93.8\pm0.67$&$86.1\pm0.7$&$88.3\pm1.2$&$94.8\pm0.6$ \\

\end{tabular}}
\label{table:table_ijb_c}
\end{table}

We have also plotted receiver operating characteristic (ROC) curve and compared with the baselines given above. The ROC curves for the IJB-A dataset are given in Fig. \ref{fig:a}. As we can clearly see from the curves, the proposed  PF-cpGAN method improves upon other methods and gives much better performance, even at a FAR of 0.001.

We have also performed the task of verification and identification using the IJB-C dataset according to the verification and the identification protocol given in the dataset. The results are provided in Table \ref{table:table_ijb_c}, showing that PF-cpGAN improve on the existing state-of-the-art methods for both verification and identification. For instance, at the false accept rate (FAR) of 0.01, the best previously-used method is PR-REM, with an average GAR of $92.1\%$. PF-cpGAN improves upon PR-REM and gives an average GAR of $93.8\%$ at the same FAR. We also observe that, for  identification, specifically, rank-1 recognition, shows an improvement over the previous best state-of-the-art method FNM \cite{FNM} by about $1.1\%$.  

\begin{figure*}[t]
\centering
\includegraphics[scale=0.16]{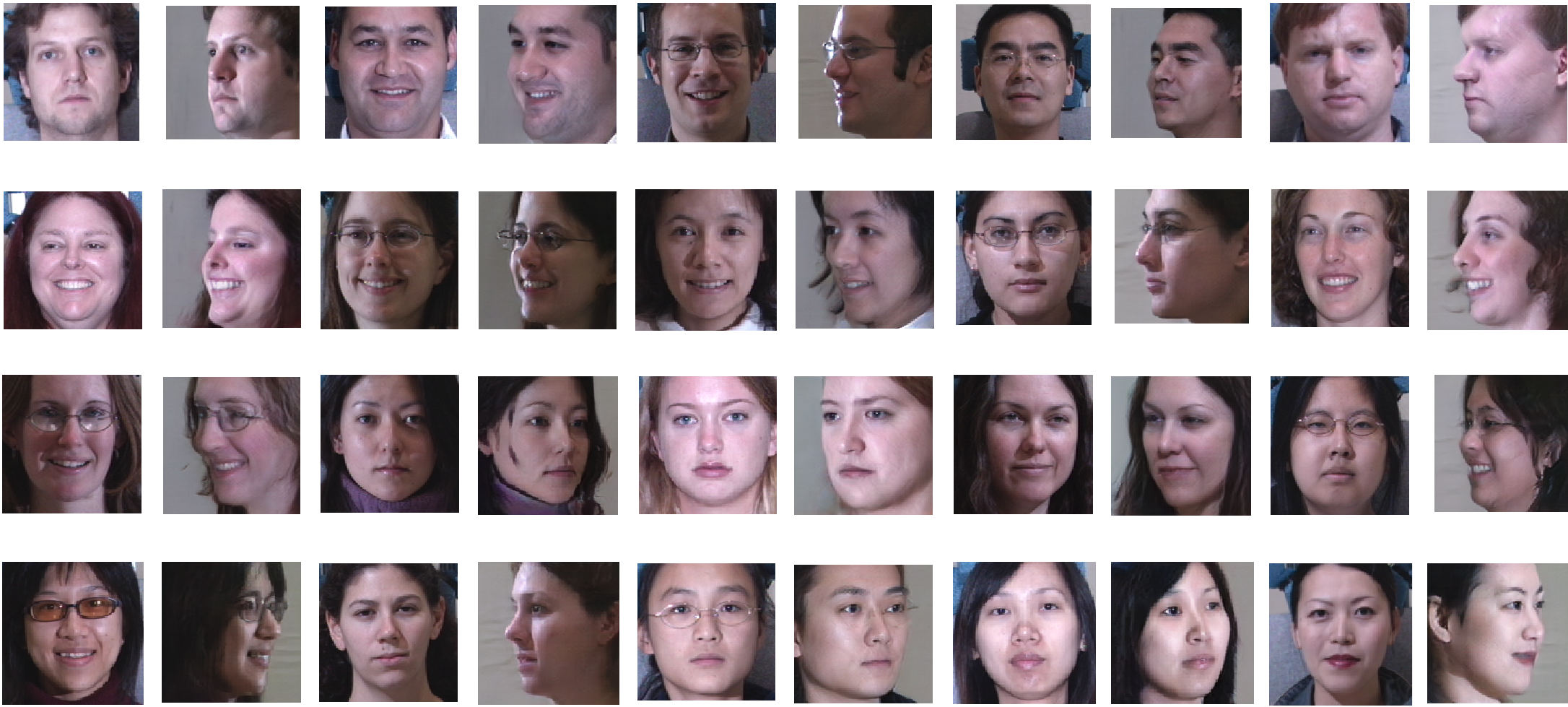}
\caption{Reconstruction of profile images at the output of the  profile U-Net generator with frontal images as input to the frontal U-Net generator. Every odd number column represents the input frontal image, and every even number column represents the output profile image. The input images belong to the CMU-MultiPIE dataset.}\label{fig:frontal_profile_images}
\vspace{-1mm}
\end{figure*}

\subsection{A Further Analysis on Influences of Face Yaw}

In addition to complete profile to frontal face recognition, we also perform a more in-depth analysis on the influence of face yaw angle on the performance of face recognition to better understand the effectiveness of  the PF-cpGAN for profile to frontal face recognition. We perform this experiment for the CMU Multi-PIE dataset \cite{CMU-PIE} under setting-1 for fair comparison with other state-of-the-art methods. As shown in Table \ref{table:multi-pie-yaw}, we achieve comparable performance with other state-of-the-art methods for different yaw angles. Under extreme pose, PF-cpGAN achieves significant
improvements (i.e., approx. $77\%$ to $88\%$ under $\pm90\degree$).

\begin{table}[t]
\centering
\caption{Rank-1 recognition rates ($\%$) across poses and illuminations under Multi-PIE Setting-1.}
\scalebox{0.75}{\begin{tabular}{c c c c c c c}
 \hline
\multicolumn{1}{c}{Method} &\multicolumn{1}{c}{$\pm90\degree$} &\multicolumn{1}{c}{$\pm75\degree$}&\multicolumn{1}{c}{$\pm60\degree$}&\multicolumn{1}{c}{$\pm45\degree$}&\multicolumn{1}{c}{$\pm30\degree$}&\multicolumn{1}{c}{$\pm15\degree$}\\ [0.5ex] 
 \hline \hline
 HPN \cite{ding2017pose} & $29.82$&$47.57$&$61.24$&$72.77$&$78.26$&$84.23$\\
 c-CNN \cite{xiong2015conditional} & $47.26$&$60.7$&$74.4$&$89.0$&$94.1$&$97.0$\\
 TP-GAN \cite{huang2017beyond} & $64.0$&$84.1$&$92.9$&$98.6$&$99.99$&$99.8$\\
 PIM \cite{zhao2018towards} & $75.0$&$91.2$&$97.7$&$98.3$&$99.4$&$99.8$\\
 CAPG-GAN \cite{hu2018pose} & $77.1$&$87.4$&$93.7$&$98.3$&$99.4$&$99.99$\\
 FNM$+$VGG-Face \cite{FNM} & $41.1$&$67.3$&$83.6$&$93.6$&$97.2$&$99.0$\\
 FNM$+$Light CNN \cite{FNM}& $55.8$&$81.3$&$93.7$&$98.2$&$99.5$&$99.9$\\
 PF-cpGAN& $88.1$&$94.2$&$97.6$&$98.9$&$99.9$&$99.9$\\

\end{tabular}}
\label{table:multi-pie-yaw}
\end{table}

For further testing on the Multi-PIE dataset under setting-1, we have also plotted ROC curves and compared with other state-of-the-art methods. The ROC curves for Multi-PIE dataset are given in Fig. \ref{fig:b}.  The curves clearly indicate that the proposed method of PF-cpGAN improves upon other methods and gives much better performance, even at FAR of 0.001.

\subsection{Reconstruction of frontal and profile images}

As noted in Sec. 1, the PF-cpGAN framework can also be used for reconstruction of frontal images by using profile images as input and vice versa. The results of reconstructing frontal images  using the profile images as input are given in Fig. \ref{fig:profile_frontal_images}, and the results of reconstructing  profile images  using the frontal images as input is given in Fig. \ref{fig:frontal_profile_images}. The reconstruction procedure for frontal images is given as follows: The profile image is given as input to the profile U-Net generator and the feature vector generated at the bottleneck of the profile generator (i.e., at the output of the encoder of the profile U-Net generator) is passed through the decoder section of the frontal  U-Net generator to reconstruct the frontal image. Similarly the reconstruction procedure for profile images is given as follows: The frontal image is given as input to the frontal U-Net generator and the feature vector generated at the bottleneck of the frontal generator (i.e., at the output of the encoder of the frontal U-Net generator) is passed through the decoder section of the profile U-Net generator to reconstruct the profile image. As we can see from  Fig. \ref{fig:profile_frontal_images}  and Fig. \ref{fig:frontal_profile_images},  the PF-cpGAN can preserve the identity and generate high-fidelity faces from an unconstrained dataset such as CMU-MultiPIE. These results show the robustness and effectiveness of PF-cpGAN for multiple use of profile to frontal matching in the latent common embedding subspace, as well as in the reconstruction of facial images.

\subsection{Ablation Study}\label{subsec:ablation}
The objective function defined in (\ref{eq:20}) contains multiple loss functions: coupling loss ($L_{cpl}$), perceptual loss ($L_{P}$), $L_2$ reconstruction loss ($L_2$), and GAN loss ($L_{GAN}$). It is important to understand the relative importance of different loss functions and the benefit of using them in our proposed method. For this experiment, we use different variations of PF-cpGAN and perform the evaluation using the IJB-A dataset. The variations are: 1) PF-cpGAN with only coupling loss and $L_2$ reconstruction loss ($L_{cpl} + L_2$); 2) PF-cpGAN with coupling loss, $L_2$ reconstruction loss, and GAN loss ($L_{cpl} + L_2 + L_{GAN}$); 3) PF-cpGAN with all the loss functions ($L_{cpl} + L_2 + L_{GAN} + L_{P}$).

We use these three variations of our framework and plot the  ROC for profile to frontal face verification using the features from the common embedding subspace. We can see from Fig. \ref{fig:ablation} that the generative adversarial loss helps improve the profile to frontal verification performance, and adding the perceptual loss (blue curve) results in an additional  performance improvement. The reason for this improvement is that using perceptual loss along with the contrastive loss leads to a more discriminative embedding subspace leading to a better face recognition performance. 
\section{Conclusion}
   We  proposed a new framework which uses a coupled GAN  for profile to frontal face recognition. The coupled GAN contains two sub-networks which project the profile and frontal images into a common embedding subspace, where the goal of each sub-network is to maximize  the  pair-wise  correlation  between profile and frontal images during the process of projection. We  thoroughly evaluated our model on several standard datasets and the results demonstrate that our model notably outperforms other state-of-the-art algorithms for profile to frontal face verification. Moreover, the improvement achieved by different losses in our proposed algorithm has been studied in an ablation study.

{\small
\bibliographystyle{ieee}
\bibliography{submission_example}
}

\end{document}